\documentclass[letterpaper,11pt]{article}
\usepackage[margin=.8in]{geometry}
\usepackage{authblk}
\usepackage{hyperref}
\usepackage[round]{natbib}
\usepackage{float}
\usepackage{subcaption}
\usepackage{graphicx}
\usepackage{amsmath,amssymb}
\usepackage[dvipsnames]{xcolor}
\hypersetup{colorlinks, 
citecolor=Blue, 
linkcolor=Blue, 
urlcolor=Violet}
\DeclareMathOperator{\E}{\mathbb{E}}
\title{Disentangled Representation Learning Using ($\beta$-)VAE and GAN}
\author[1]{Mohammad Haghir Ebrahimabadi}
\affil[1]{\small Department of Computer Science, Indiana University, Bloomington, IN 47408, USA}
\date{}
\begin{document}
\maketitle

\begin{abstract}
Given a dataset of images containing different objects with different features such as shape, size, rotation, and x-y position; and a Variational Autoencoder (VAE); creating a disentangled encoding of these features in the hidden space vector of the VAE was the task of interest in this paper. The dSprite dataset provided the desired features for the required experiments in this research. After training the VAE combined with a Generative Adversarial Network (GAN), each dimension of the hidden vector was disrupted to explore the disentanglement in each dimension. Note that the GAN was used to improve the quality of output image reconstruction.
\end{abstract}
\section{Introduction}

Truly understanding a data might require identifying its generative factors. A concept that is more formally known as disentanglement. Classical approaches such as Principal Component Analysis (PCA) \citep{zietlow2021demystifying} has been developed for this purpose using linear algebra. In addition, VAE \citep{kingma2013auto} is a learning-based architecture that aims to represent the data in its disentangled latent space. In other words, VAEs were developed for learning a latent manifold that its axes align with independent generative factors of the data. \citet{zietlow2021demystifying} argued that VAEs recover the nonlinear principal components of the data. In addition $\beta$-VAEs \citep{higgins2016beta} are a modified version of VAEs that when $\beta > 1$, weigh in more for disentanglement by sacrificing reconstruction quality. In this project, the goal is to explore the capacities of $\beta$-VAEs for learning a disentanglement representation, specifically to what degree the position of a moving object in the input frames can be encoded in the latent space.


\section{Background and Related Work}\label{sec:back}
Learning the posterior distribution of continuous latent variables in probabilistic models is intractable. \citet{kingma2013auto} proposed a Variational Bayesian (VB) approach for approximating this distribution that can be learned using stochastic gradient descent. This approach can be used in different settings where latent variables of a model required to be learned e.g. in supervised models with latent variable and learning complicated noise distributions. One example of using this approach for noise identification and removal is presented in \citep{wan2020old}. In this work VB is used in an encoder-decoder setting which is known as VAE. In VAEs, the goal is learning the latent variables for input reconstruction. The loss in this setting is consisted of a reconstruction loss and a disentanglement loss. One variation of VAE is $\beta$-VAE \citep{higgins2016beta, burgess2018understanding} where the second term in the loss function of VAEs can be controlled using a parameter $\beta$. This parameter can be used to establish a trade-off between the reconstruction accuracy and disentanglement of the learned representations in the latent space. Thus, there are two measures that are of interest in this setting: (I) the amount of disentanglement (II) the reconstruction output accuracy. Factor-VAE Metric (FVM) \citep{kim2018disentangling} and Mutual Information Gap (MIG) \citep{chen2018isolating} are developed for quantifying disentanglement. In addition, Frechet Inception Distance (FID) \citep{heusel2017gans} was developed for measuring the generated output quality. Note that, at some point in the project, I also used a GAN module. Details of relevant previous works on GAN are mentioned in Section \ref{sec:method}.

\section{Methods}\label{sec:method}

$\beta$-VAEs ($\beta > 1$ ) showed to have higher performance in disentanglement representation learning and generation quality compared to their peers such as VAEs ($\beta = 1$), InfoGAN \citep{chen2016infogan}, and DC-IGN \citep{kulkarni2015deep} on the dSprite dataset \citep{dsprites17}. In this project, first I aimed to reproduce the output of $\beta$-VAEs for various learning configurations such as the number of latent dimension, $\beta$ parameter value, and the learning rate. This phase of the project required understanding of $\beta$-VAEs loss function and auto-encoders structure, and their implementation. Essentially, the objective function in $\beta$-VAE is to optimize a modified lower bound of the marginal likelihood as follows:
\begin{equation}
 \E_{x \sim p(x)}[\log p(x)] \geq \E_{z \sim q_{\phi}(z | x)}[\log p_{\theta}(x | z)] - \beta D_{KL}(q_{\phi}(z|x) || p(z))  
\end{equation}

where $x$ is a data point and the first term aims for a higher generation quality and the KL divergence term \citep{burgess2018understanding} forces the posterior to be closer to the prior $p(z)$ which results in a more disentangled representation. Note that higher values of $\beta$ sacrifices the generation quality in favor of a more disentangled representation in latent space.

Throughout the first phase of the project, I observed that even by setting the value of $\beta$ to values $<1$, the generation quality was far from the ground truth. This was observed even though the experiments were performed on a synthetic dataset in a controlled manner and without complications of a real-world dataset. This poor generation quality might arise from the fact that; (I) some factors of data might actually be at least partially dependent, so our simplifying assumption does not fully hold (II) the generator is usually a simple decoder and not capable of rendering complex patterns in output. To alleviate this problem, \citep{lee2020high} proposed ID-GAN that feeds the latent space of VAE to a GAN \citep{goodfellow2014generative}. The GAN module is employed in order to generate an output with high fidelity. This approach combines the strengths of the two modules; disentanglement representations form VAEs and high-fidelity synthesis of GANs. The disentangled factors acquired by the VAE module form the distilled information that will be the input to the GAN module.

A GAN module consists of a generator $G$ and a discriminator $D$. The input to the generator is a noise variable $z$, and it aims to generate a fake sample from $z$ that maximizes the probability of the discriminator to make a mistake in identifying the true sample from the fake sample. The objective function V is as follows:
\begin{equation}
    \min_{G} \max_{D} V(D, G) = \E_{x \sim p_{data}(x)}[\log D(x)] + \E_{z \sim p_z(z)}[\log (1 - D(G(z)))]
\end{equation}
 where $x$ is a data point as before \citep{goodfellow2014generative}. Note that, here is no constraint on $z$ in this formulation. In InfoGAN \citep{chen2016infogan}, this noise vector was decomposed to two parts; (I) a noise vector $z$ (II) a latent code $c$ that aims to represent the salient semantic features of the data distribution. Essentially, in InfoGAN a regularization term was added to the objective function for maximizing the mutual information between the latent code $c$ and the generator distribution $G(z, c)$ \citep{chen2016infogan}. Thus, the InfoGAN objective function is:
 \begin{equation}
     \min_{G} \max_{D} V(D, G) - \lambda I(c, G(z, c))
 \end{equation}

The mutual information term $I(c, G(z, c))$ includes a posterior $p(c|x)$ (similar to VAEs), which cannot be optimized directly. Therefore, a lower bound can be calculated for this term by introducing an approximate posterior $q_{\phi}(c|x)$ for $p(c|x)$. Using this assumption, the lower bound is as follows:
\begin{equation}
    \E_{z \sim p(z)}[D_{KL}(p(c)||q_{\phi}(c|G(z,c)))]
\end{equation}

\begin{figure}[H]
   \centering
   \includegraphics[width=13
   cm]{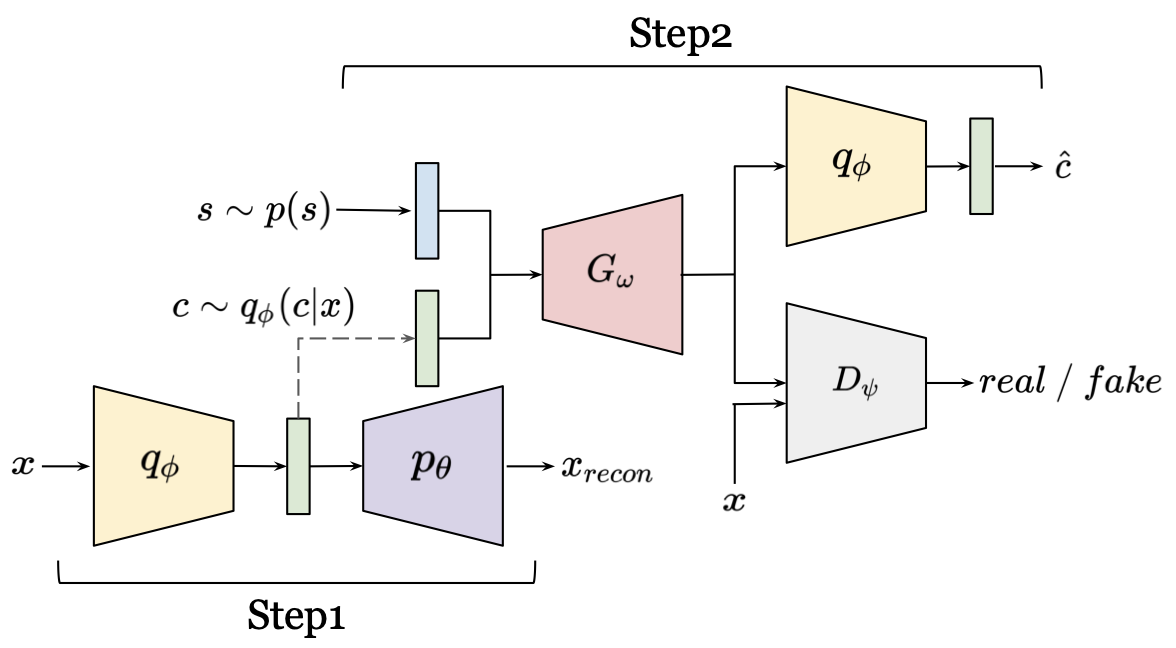}
   \caption{The ID-GAN network. The $\beta$-VAE model is trained in Step1. Then the latent space of the trained $\beta$-VAE model is concatenated with the noise vector as input to the generator, and the GAN module is trained. The variable $s$ in this figure is written as $z$ in the equations above. The schematic was taken from \citep{lee2020high}.}
   \label{fig:idgan_net}
\end{figure}

Note that in this formulation, the latent code $c$ is computed as part of the whole GAN module which can degrade the disentanglement performance. In this project, I used the formulation developed in ID-GAN \citep{lee2020high} that learns the latent code separately using $\beta$-VAE. Using this formulation, I can use the latent space of the $\beta$-VAE models I trained for the first phase. Note that in the ID-GAN formulation, the regularization term is as follows:

\begin{equation}
    \beta R_{VAE}(q) + \lambda R_{ID}(G)
\end{equation}\label{eq:reg_idgan}
\begin{equation}
     R_{VAE}(q) = \E_{x \sim p(x)}[D_{KL}(q_{\phi}(c|x)||p(c))]
\end{equation}

\begin{equation}
     R_{ID}(q) = \E_{s \sim p(s)}[D_{KL}(q_{\phi}(c)||q_{\phi}(c|G(z, c)))]
\end{equation}
The architecture of the ID-GAN network is shown in Figure \ref{fig:idgan_net}. In Step1, the $\beta$-VAE model is trained. The latent code of the trained $\beta$-VAE is concatenated with the input noise vector to the generator for training in Step2. The variable $s$ in Figure \ref{fig:idgan_net} is written as $z$ in the equations above.

\newpage
\section{Experiments}
In this section, I explain the properties of the dSprite dataset. In addition, I describe the experiments that have been done and the obtained results. Note that the implementations have been done in PyTorch.

\section{Dataset}
For the experiments, dSprite dataset has been used. This synthetic dataset consist of $737,280$ binary 2D shapes. The dataset contains all combinations of $3$ different shapes (oval, heart and square) with $4$ other attributes: (i) $32$ values for position X (ii) $32$ values for position Y (iii) $6$ values for scale (iv) $40$ values for rotation. The images in this dataset are of $64 \times 64$ resolution. For each of these attributes we have equal number of labels as the number of distinct values. Using these labels a subset of the dataset can be selected. 

\subsection{Results}

\begin{figure}[H]
   \centering
   \includegraphics[width=13
   cm]{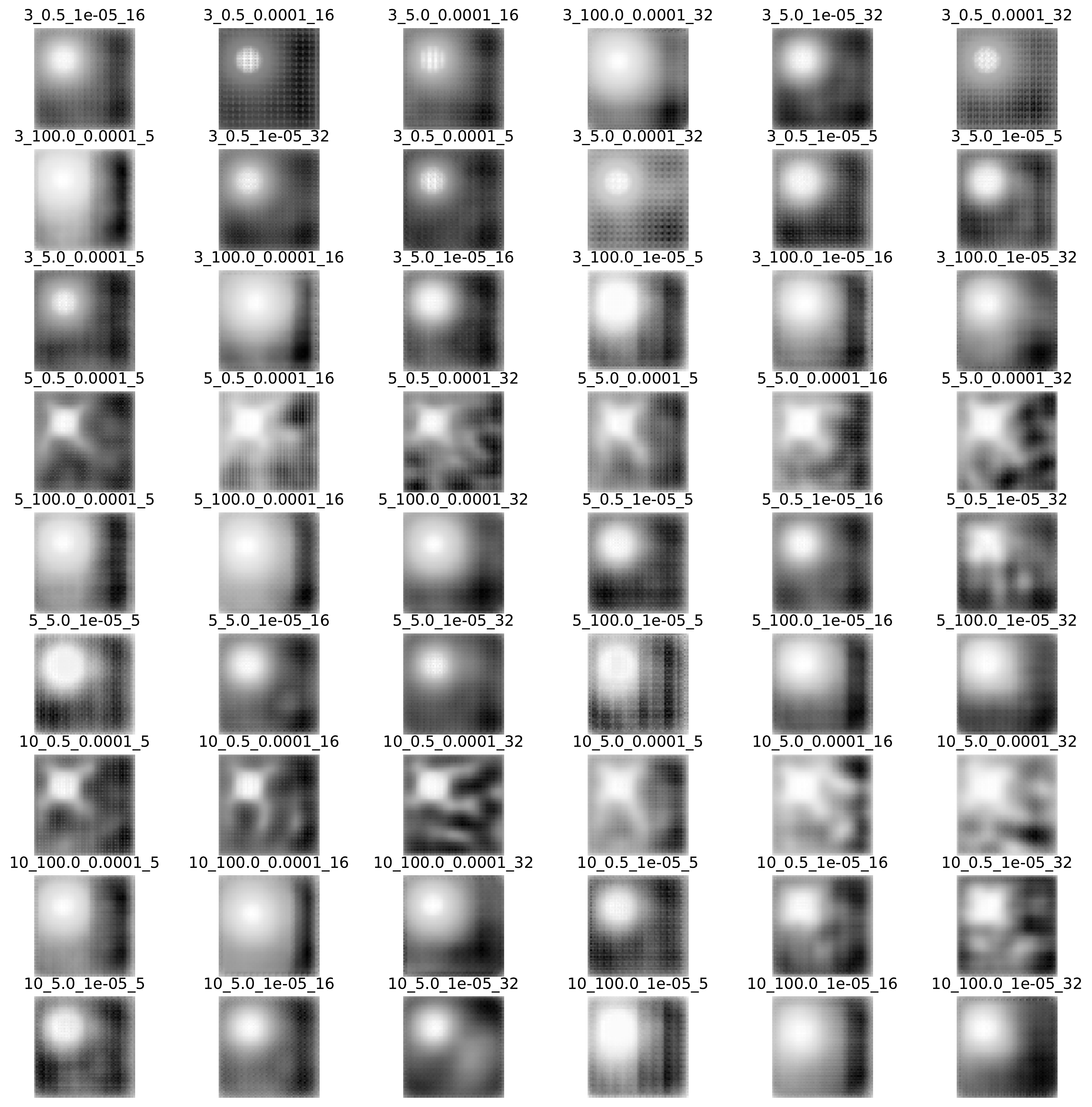}
   \caption{The reconstructed frames obtained by feeding a frame to a trained $\beta$-VAE with the setting in the title. The numbers in the title of each output are latent dimension, $\beta$ value, learning rate, and threshold for excluding some of the x-axis positions from training data, respectively.}
   \label{fig:bvae_ds}
\end{figure}

In the first phase of the project I trained several $\beta$-VAE models with different settings for the number of dimension of latent space, value of $\beta$, learning rate, and the number of epochs. Figure \ref{fig:bvae_ds} shows the decoder/generation output of a frame for models trained with different settings. The numbers in the title of each output are latent dimension, $\beta$ value, learning rate, and threshold for excluding some of the x-axis positions from training data, respectively. I chose the settings to be all of the combinations of $|z| \in \{3, 5, 10\}$, $\beta \in \{0.5, 5, 100\}$, learning rate $\in \{1\mathrm{e}{-4}, 1\mathrm{e}{-5}\}$, and position threshold $\in \{5, 16, 32\}$. A position threshold $t$ means that only the samples with x-axis position label $\leq t$ will be considered in training. I used this threshold parameter since I wanted to exclude some of the positions from training, and see if the learned latent space can be generalized to produce a sample in an unseen position during training. This hypothesis can be evaluated by traversing the latent space in a systematic manner (I have done this for the ID-GAN that I talked about later in this section).
As can be seen in Figure \ref{fig:bvae_ds}, all of the models could capture the position of the object in the frame, i.e. the circle with high intensity, however the generation quality is far from the input frame. Note that, all of the models were trained for $100$ epochs and the batch size for all of them was $256$. It is worth mentioning that training the same experiments for $1000$ epochs did not change the output, so I included the ones for $100$ epochs here.

Notice that the generation quality in Figure \ref{fig:bvae_ds} is better than others for some of the settings for example when $|z| = 3, \: \beta = 0.5$, learning rate (lr) is $0.0001$ and the x-axis threshold (tr) is $16$. However, still the shape boundaries are not sharp. To improve the generation quality of this model, I chose four settings from Figure \ref{fig:bvae_ds} and used the latent code of their model as input $c$ for Step2 of training the ID-GAN. In the following I discuss the output of the ID-GAN for each setting.

\vspace{0.4cm}

For all of the settings, Step2 in Figure \ref{fig:idgan_net} is trained for $100$ epochs. One way for evaluating the disentanglement is by traversing the latent code and evaluate the output qualitatively \citep{kim2018disentangling}. For example traversing one of the latent dimensions in a range and keeping the rest fixed might change the position of the object in different frames. This is a sign that this dimension is representing the position of the object. Furthermore, in all of the settings the generation quality is much higher than $\beta$-VAE alone, i.e. the boundaries of shapes are sharp, and the background is more clear.

\begin{figure}[H]
   \centering
   \includegraphics[width=16
   cm]{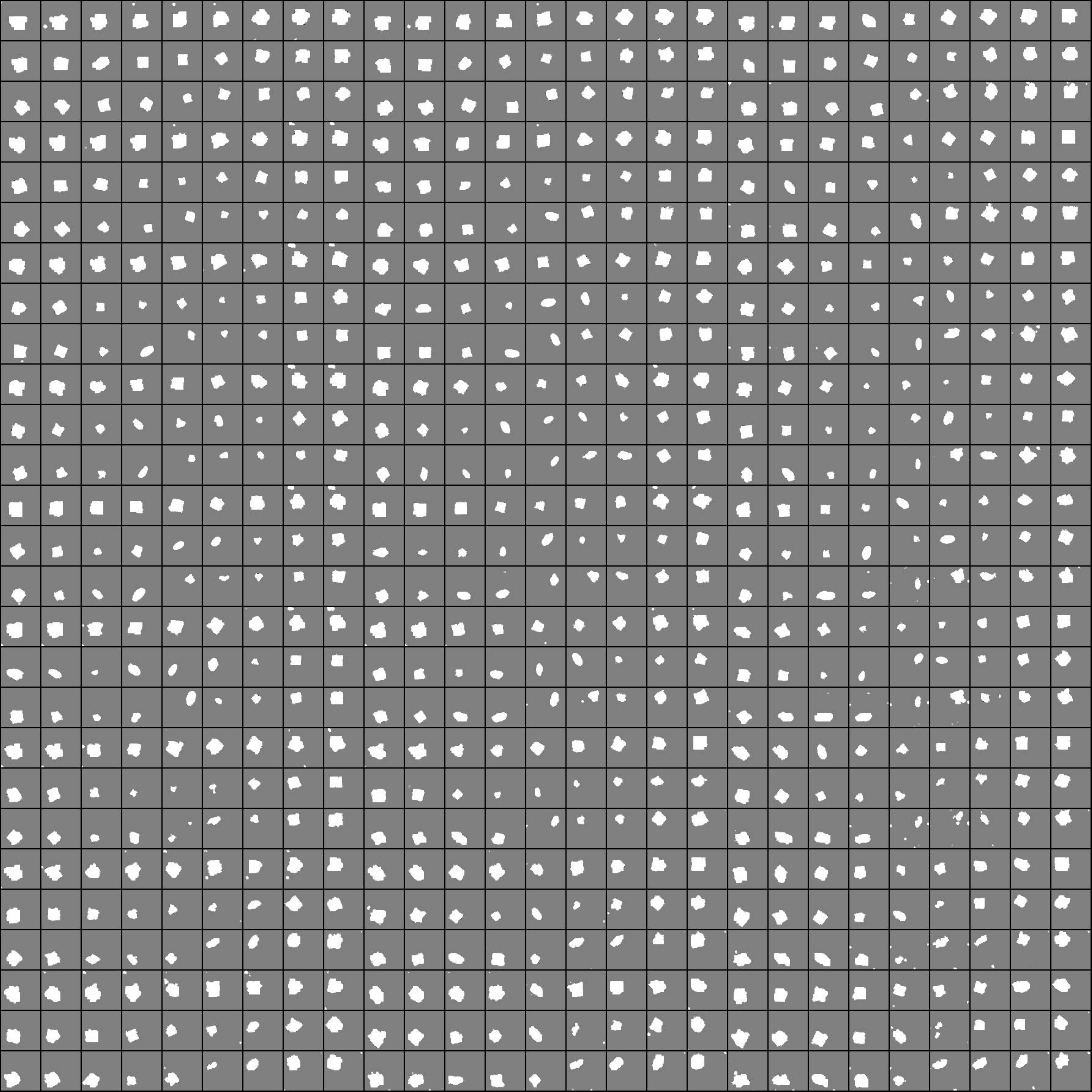}
   \caption{The ID-GAN output by traversing the latent code in the range $[-2, 2]$. The settings for $\beta$-VAE: $|z| = 3, \: \beta = 0.5, \: lr = 0.0001, \: tr = 16$.}
   \label{fig:vae_gan_3_0.5}
\end{figure}

\paragraph{The first setting.} $|z| = 3, \: \beta = 0.5, \: lr = 0.0001, \: tr = 16$. I traversed the latent code in the range $[-2, 2]$ with steps $0.5$. That means I used all the combinations of nine numbers for different dimensions (the combinations are generated using three nested for loop). Figure \ref{fig:vae_gan_3_0.5} shows the output for these latent code values. There are $27$ columns in this figure which means for the outputs in each row while two of the dimensions are fixed, the other one can take three consecutive values. Looking at the first column of this figure, it seems the first latent dimension is controlling the vertical position of the shape (this dimension is changing only vertically). Although, following the rows we see that the position, shape, scale, and rotation of the shapes are changing periodically and that can be a sign that not just one dimension is controlling one properties. Overall, looking at these output, different patterns can be discovered; for example looking at the few last rows I do not see any small scale shapes which can indicate that a combination of larger values for latent code can prevent from generating small shapes. Also, some shapes are generated at the second half of x-axis that the $\beta$-VAE network did not see during training.  

\paragraph{The second setting.} $|z| = 3, \: \beta = 5.0, \: lr = 0.0001, \: tr = 16$. In this setting, the $\beta$ value was higher than the first setting (also $> 1$) with everything else unchanged. I see more disentanglement at least for scale. However, higher value of $\beta$ degraded the generation quality. Similar to the first setting, it seems still a combination of latent dimensions can change the properties of generated output rather than one dimension.
\begin{figure}[H]
   \centering
   \includegraphics[width=16
   cm]{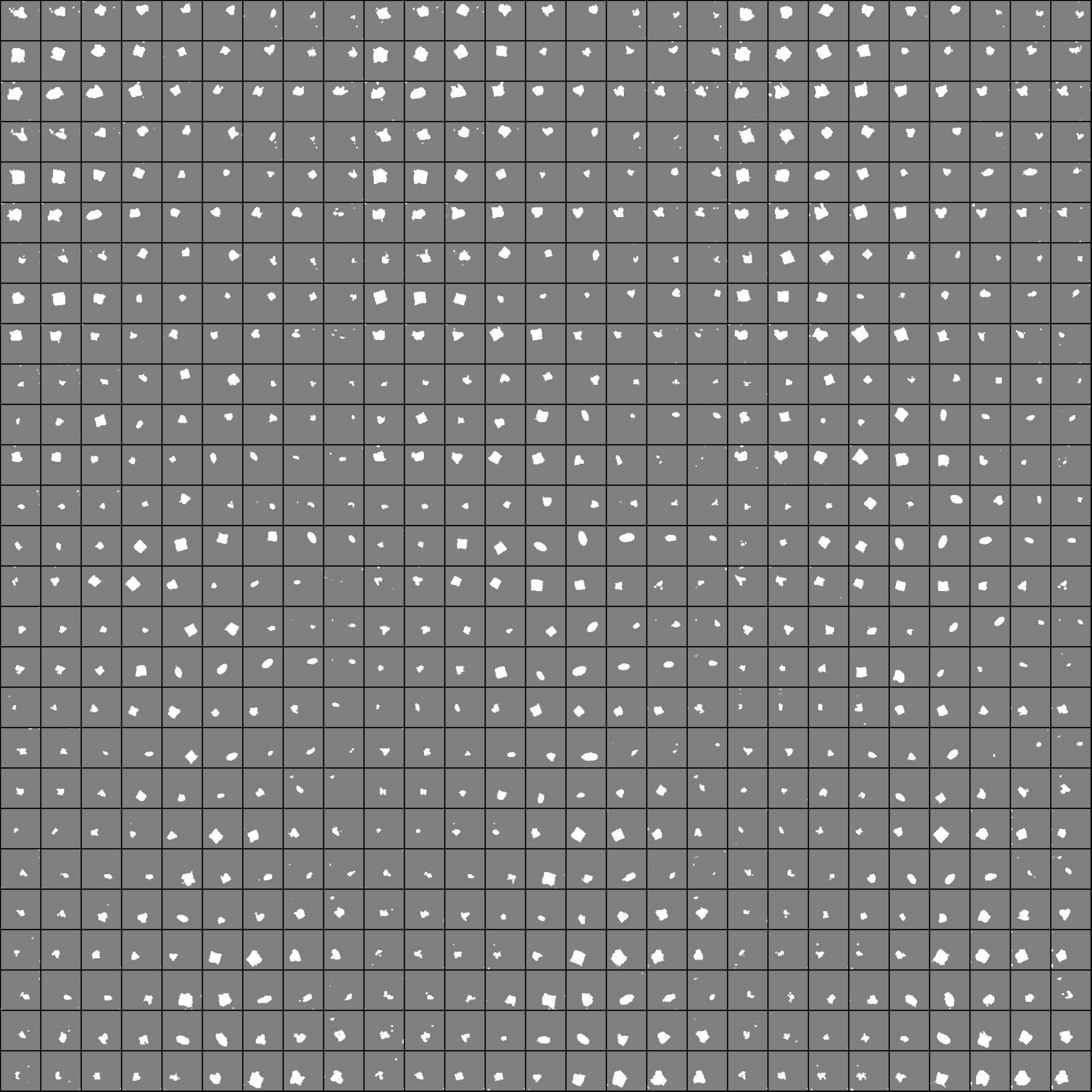}
   \caption{The ID-GAN output by traversing the latent code in the range $[-2, 2]$. The settings for $\beta$-VAE: $|z| = 3, \: \beta = 5.0, \: lr = 0.0001, \: tr = 16$.}
   \label{fig:vae_gan_3_5.0}
\end{figure}

\paragraph{The thrid setting.} $|z| = 5, \: \beta = 0.5, \: lr = 0.0001, \: tr = 16$. I traversed the latent code in the range $[-2, 2)$ with steps $1$. So, the latent code can take any combination of five numbers. In this setting, I see that the scale is changing less than previous settings that can be a sign of higher disentanglement. Also, it seems the y-axis position of shapes is only changing on certain values for a specific dimension; shapes were generated at the top of the frame and periodically their position were changed to the bottom of the frame.

\begin{figure}[H]
   \centering
   \includegraphics[width=16
   cm]{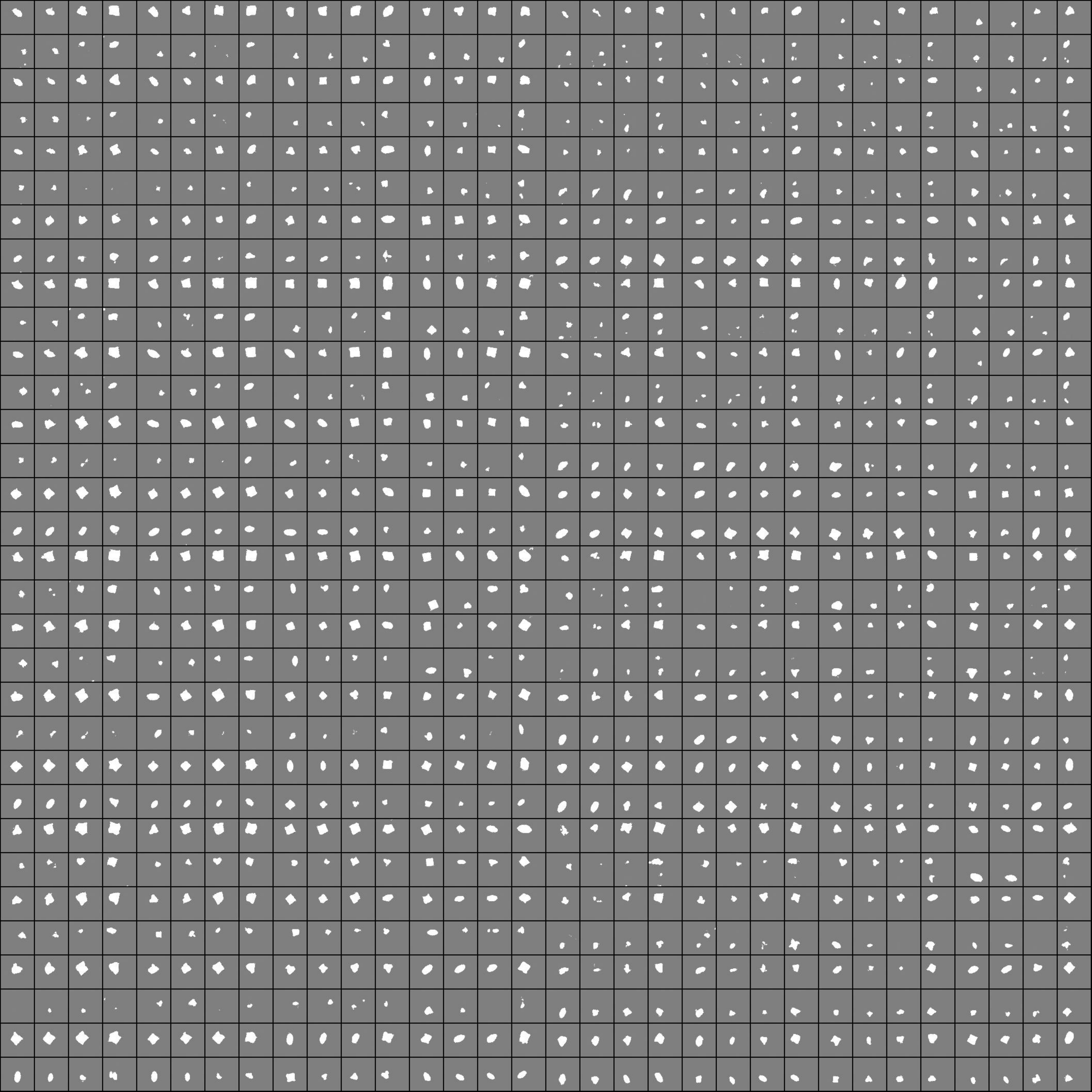}
   \caption{The ID-GAN output by traversing the latent code in the range $[-2, 2)$. The settings for $\beta$-VAE: $|z| = 5, \: \beta = 0.5, \: lr = 0.0001, \: tr = 16$.}
   \label{fig:vae_gan_5_0.5}
\end{figure}

\begin{figure}[H]
   \centering
   \includegraphics[width=16
   cm]{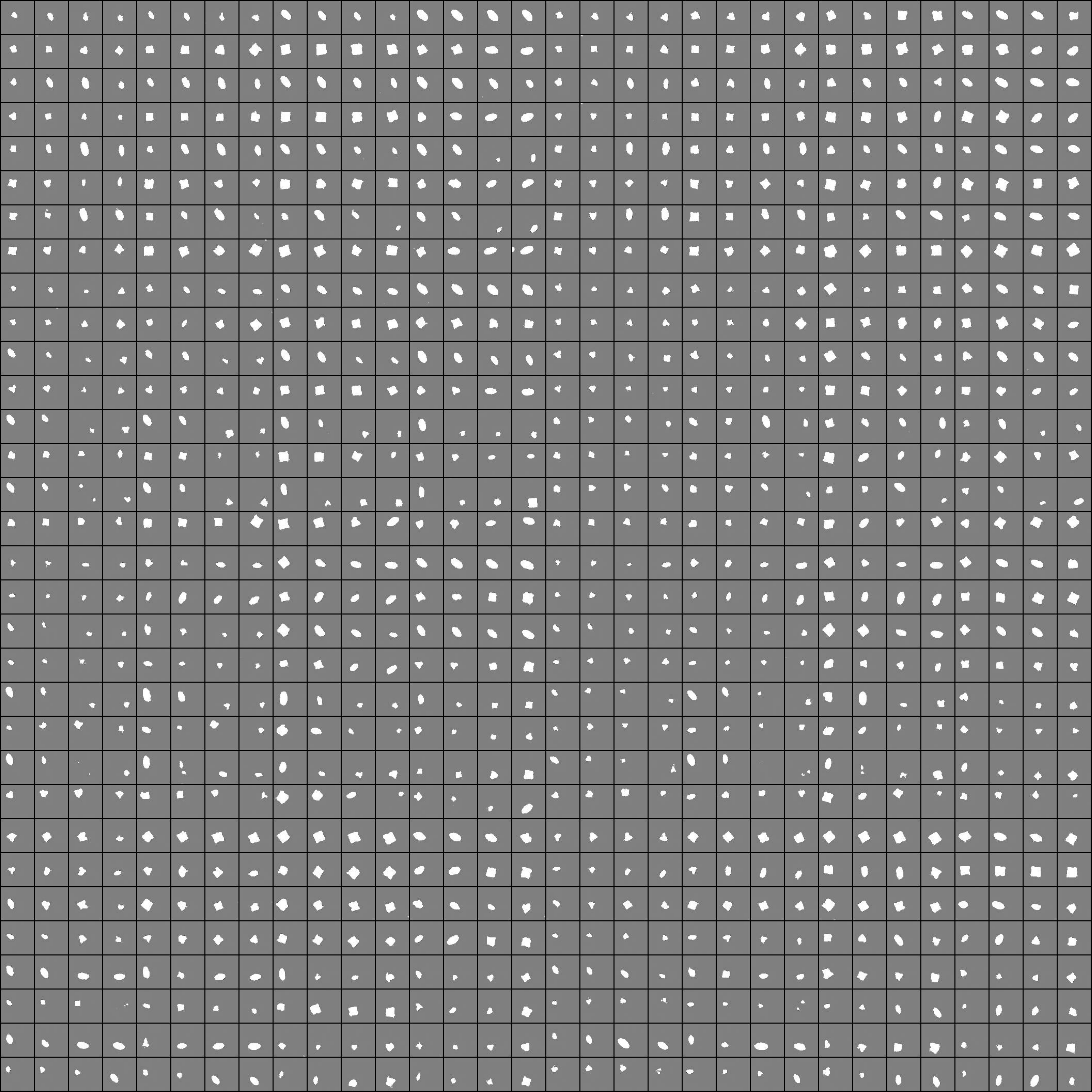}
   \caption{The ID-GAN output by traversing the latent code in the range $[-2, 2)$. The settings for $\beta$-VAE: $|z| = 5, \: \beta = 5.0, \: lr = 0.0001, \: tr = 16$.}
   \label{fig:vae_gan_5_5.0}
\end{figure}

\paragraph{The fourth setting.} $|z| = 5, \: \beta = 5.0, \: lr = 0.0001, \: tr = 16$. Similar to the previous setting, I see scales are limited in Figure \ref{fig:vae_gan_5_5.0} which can be a sign of disentanglement. Also, the generation quality is higher in this case. Also, shapes and y-axis positions are periodically changing.

\section{Discussion}
In this paper I explored the disentanglement and generation performance of $\beta$-VAEs. At this stage, I observed that generation quality is not as good that I can even investigate the disentanglement performance. So, I used an architecture called ID-GAN to improve the generation quality. The output from ID-GAN has a much higher generation quality, also I observed some degrees of disentanglement. I evaluated the performance using latent code traversal which can be subjective. There are some metrics developed for quantifying both generation quality and disentanglement that I mentioned some of them in Section \ref{sec:back}. A caveat about these metrics is that the ground truth disentangled representation of the dataset is needed for being able to calculate them.

\section{Conclusion}

I evaluated the performance of $\beta$-VAEs for disentanglement and generation. Furthermore, I looked for ways to improve the $\beta$-VAE performance which led me to some works that discuss the correspondence of $\beta$-VAE with PCA \citep{zietlow2021demystifying}, how to alleviate the generation output of VAEs \citep{lee2020high}, also, is $\beta$-VAE prioritizing in retaining information (or any information is equally good) \citep{fertig2018beta}. I also realized it is possible to improve the generation quality of $\beta$-VAEs by using a GAN module.

Quantifying the disentanglement and understanding the number of required dimensions for encoding a feature could be a topic of interest for future research. Although using a GAN helped with having a closer reconstructed output image to the input, that might have a reversed effect on obtaining disentangled features. In this regard, simplifying the network for the main task could result in a more precise answer.
\newpage
\bibliographystyle{agsm}
\bibliography{references}
\end{document}